\pdfoutput=1

\documentclass[11pt]{article}

\usepackage[final]{acl}

\makeatletter
\let\@LN@orig@makecol\@makecol       
\def\@LN@makecol{%
  \@LN@orig@makecol                   
  \setbox\@outputbox\vbox{
    \boxmaxdepth\@maxdepth
    \protected@write\@auxout{}{
      \string\@LN@col{\if@firstcolumn1\else2\fi}%
    }%
    \box\@outputbox                   
  }%
}
\AtBeginDocument{\let\@makecol\@LN@makecol}  
\makeatother

\usepackage{times}
\usepackage{latexsym}

\usepackage{amsmath}
\usepackage{amssymb}
\usepackage{amsfonts}

\usepackage{algorithm}
\usepackage{algorithmic}

\usepackage{tikz}
\usetikzlibrary{arrows.meta, positioning, calc, fit, backgrounds}

\usepackage{booktabs}
\usepackage{placeins}
\usepackage{multirow}
\usepackage[table]{xcolor}

\usepackage{subcaption}

\usepackage{graphicx}

\usepackage{listings}
\makeatletter
\patchcmd{\lst@Init}{\begingroup}{\par\begingroup\nolinenumbers}{}{}
\makeatother

\usepackage[T1]{fontenc}

\usepackage[utf8]{inputenc}

\usepackage{microtype}

\usepackage{inconsolata}

\title{Faithful Autoformalization via Roundtrip Verification and Repair}

\author{Daneshvar Amrollahi$^{\dagger}$ \\
  \normalfont\normalsize \texttt{daneshvar@cs.stanford.edu}
  \And
  Jerry Lopez$^{\ddagger}$ \\
  \normalfont\normalsize \texttt{lopezjerry721@gmail.com}
  \And
  Clark Barrett$^{\dagger}$ \\
  \normalfont\normalsize \texttt{barrett@cs.stanford.edu}
  \AND
  \normalfont\normalsize $^{\dagger}$Stanford University \quad $^{\ddagger}$Independent Researcher}

\begin{document}
\maketitle
\begin{abstract}
When an LLM formalizes natural language, how do we know the output is faithful?
We propose a roundtrip verification approach which does not require ground-truth annotations: formalize a statement, translate the result back to natural language, re-formalize, and use a formal tool to check logical equivalence.
When the two formalizations agree, this provides evidence of a faithful formalization.
When they disagree, a stage-level diagnosis localizes the error to a specific translation step, and a scoped repair operator attempts to correct that step.
We evaluate the framework on two statutory domains (the Texas Transportation Code and the Texas Parks and Wildlife Code) using two LLMs (Claude Opus~4.6 and GPT-5.2) with three repair baselines.
Diagnosis-guided scoped repair is the most effective method, with effectiveness contingent on the reliability of the diagnosis function.
Across both domains and both models, under our full repair system, rules that fail the equivalence check show 1.4$\times$-2.5$\times$ more natural language inference (NLI) drift than rules that pass it.
\end{abstract}

\section{Introduction}
\label{sec:intro}

Autoformalization, the automatic translation of natural-language statements into formal, machine-checkable representations, is a rapidly growing area of NLP research \citep{wu2022autoformalization}.
Yet a fundamental question remains largely unaddressed: \emph{when an LLM produces a formalization, how do we know it faithfully captures the meaning of the original text?}
Existing evaluations rely on syntactic validity or downstream task success,
neither of which directly tests semantic fidelity. Furthermore, when errors are found, repair is limited to end-to-end regeneration with no diagnosis of where meaning was lost (\S\ref{sec:related}).

We propose a framework that addresses both problems (verification and repair) without requiring ground-truth formalizations.
The key idea is simple (Figure~\ref{fig:roundtrip}): given a natural-language
specification $x$, we formalize it ($T_1$), translate the result back to
natural language ($T_2$), and then re-formalize the result ($T_3$).
We call each of these translation steps a \emph{stage}.
If the pipeline preserves meaning, the initial and final formalizations should be semantically equivalent.
When they disagree, the point of divergence localizes the error to a specific
stage, enabling both valuable diagnostic information as well as the possibility
of \emph{scoped repair}: attempting to correct only the faulty component rather than regenerating everything.
The framework is agnostic to the choice of the formalism: it applies whenever
a target formalism admits an equivalence check, e.g., automated reasoning tools for
logical formulas, type-checkers for proof assistants, test suites for code, etc.

We evaluate the framework on two statutory domains: 150 rules from the Texas Transportation Code and 77 rules from the Texas Parks and Wildlife Code.
We use two LLMs, Claude Opus~4.6 (Anthropic) and GPT-5.2 (OpenAI).
The target formalism is based on many-sorted first-order logic, with equivalence checked by an SMT solver~\cite{BSST21}.
We compare our diagnosis-guided scoped repair against three baselines: no repair, repair with random stage selection, and full-pipeline regeneration.
The results yield two principal findings:
\begin{enumerate}
\item \textbf{Formal equivalence predicts semantic faithfulness.}
Across both domains and both models, under our full repair system, rules that fail the equivalence check show 1.4$\times$-2.5$\times$ more NLI drift than rules that pass it.
\item \textbf{Diagnosis-guided scoped repair achieves the highest verified-equivalence rate at the lowest cost when the diagnosis step is reliable.}
With Claude performing the pipeline, repair, and diagnosis, scoped repair leads on both domains.
With GPT in those roles, diagnosis is unreliable, causing scoped repair to lose its lead.
Replacing only the diagnosis step with Claude (keeping GPT for the pipeline and repair) restores scoped repair to a clear lead at lower cost on both domains.
\end{enumerate}

Our contributions are:
(i)~A roundtrip verification framework that signals faithfulness of autoformalization without requiring ground-truth formalizations, instantiated with SMT in our experiments and applicable in principle to any target formalism with an equivalence check.
(ii)~A stage-level diagnosis and scoped repair procedure that localizes errors to individual translation steps. Repair is driven by formal equivalence alone. A bidirectional NLI signal serves as an independent post-hoc semantic check.
(iii)~An empirical study across two statutory domains and two model families that identifies the diagnosis step (not the pipeline or repair operators) as the bottleneck for scoped repair.

\begin{figure}[t]
\centering
\setlength{\fboxsep}{6pt}\setlength{\fboxrule}{0.4pt}%
\fcolorbox{gray!40}{white}{%
\begin{tikzpicture}[
    node distance=1.8cm and 3.0cm,
    box/.style={draw, rounded corners=3pt, minimum width=1.8cm, minimum height=0.85cm, align=center, font=\small},
    nlbox/.style={box, fill=blue!8},
    fmlbox/.style={box, fill=orange!10},
    arrow/.style={-{Stealth[length=5pt]}, thick},
    lbl/.style={font=\footnotesize, fill=white, inner sep=1.5pt},
  ]
  \node[nlbox]  (x)  {$x$};
  \node[fmlbox, right=of x] (y) {$y^{\mathrm{orig}}$};
  \node[nlbox,  below=of y] (xp) {$x'$};
  \node[fmlbox, below=of x] (yrt) {$y^{\mathrm{rt}}$};

  \draw[arrow] (x)  -- node[lbl, above] {$T_1$} (y);
  \draw[arrow] (y)  -- node[lbl, right] {$T_2$} (xp);
  \draw[arrow] (xp) -- node[lbl, below] {$T_3$} (yrt);

  \draw[{Stealth[length=5pt]}-{Stealth[length=5pt]}, dashed, thick, red!70!black]
    (y.south west) -- node[lbl, left, text=red!70!black, align=center]
    {$\equiv_{\Sigma}$\,?} (yrt.north east);
\end{tikzpicture}}%
\caption{Roundtrip loop.  An input $x$ is formalized ($T_1$), reconstructed
back to natural language ($T_2$), and re-formalized ($T_3$).  The diagonal
compares $y^{\mathrm{orig}}$ and $y^{\mathrm{rt}}$ for semantic equivalence under $\Sigma$.}
\label{fig:roundtrip}
\end{figure}

\section{Related Work}
\label{sec:related}

\paragraph{Autoformalization.}
Autoformalization \citep{wu2022autoformalization} has produced NL-formal benchmarks for proof assistants
\citep{zheng2022minif2f,azerbayev2023proofnet,ying2024leanworkbook,Gao2024Herald,jiang2023mma} and
methods that use prover or compiler feedback to improve formalization quality 
\citep{jiang2023dsp,lu2024processdriven,yang2023leandojo,murphy2024autoformalizing}.
Both lines score a formalization by whether it compiles or whether a downstream proof succeeds.
Concurrent work formalizes non-mathematical text \citep{manas2024tr2mtl}.
Formal specifications have also been derived from natural-language system requirements \citep{cosler2023nl2spec}.
These two efforts target temporal logic, whereas we target many-sorted first-order logic.
Across these settings, fidelity to the source text is either untested or left to a human.

\paragraph{Verification-integrated repair.}
Several systems couple verification with repair: Clover checks mutual
consistency among LLM-generated code, docstrings, and formal annotations via a
Dafny verifier \citep{sun2023clover}, but it assumes the existence of the
formalization from the start.  ProofBridge iteratively repairs Lean proofs from type-checker 
feedback \citep{jana2025proofbridge}. DeepSeek-Prover-V1.5 uses RL with proof-assistant 
rewards \citep{xin2025deepseekproverv}. Self-debugging repairs code from execution results \citep{chen2023selfdebugging}.
In a related approach, LLMs translate a problem into a formal representation and a solver performs the
reasoning \citep{pan-etal-2023-logic, olausson-etal-2023-linc, 3666122.3668096}, with Logic-LM
feeding solver error messages back to revise the formalization.
Where feedback drives repair, the system regenerates the artifact it checked.
To our knowledge, none localizes the error to a particular stage of a multi-step translation pipeline.
A parallel self-refinement line \citep{3666122.3668141, 10.5555/3666122.3666499, 10.1162/tacl_a_00660} 
forgoes external verification entirely and uses the LLM as its own feedback source.

\paragraph{Faithfulness estimation.}
Faithfulness in natural language generation (NLG) asks whether generated text is supported by its source, 
and it is widely studied \citep{maynez2020faithfulnessfactualityabstractivesummarization, honovich-etal-2022-true-evaluating}.
Support between two natural-language texts has no decision procedure, so faithfulness is estimated by a learned metric rather than decided.
Our equivalence check is decided by a solver, and we use a learned metric only as an auxiliary diagnostic (\S\ref{sec:nli}).

\paragraph{Roundtrip agreement.}
In machine translation, roundtrip agreement between a sentence and its back-translation estimates quality without
a reference \citep{moon-etal-2020-revisiting, zhuo-etal-2023-rethinking}.
Roundtrip correctness has been used as an evaluation signal in code generation
when checking equivalence via test suites \citep{allamanis2024rtc}.
Closest to our setting, \citet{karia2025autoeval} assess LLMs with a roundtrip between formal 
expressions and natural language.
Their roundtrip starts from a formal expression the system generates from a grammar, so that 
expression is its own reference.
The framework therefore measures how well LLMs translate, rather than producing a trustworthy 
formalization of a given input.
Our input is a human-written rule with no formal reference.

\paragraph{Semantic consistency checks.}
Agreement among model outputs is also used as evidence of correctness, as in hallucination detection 
by self-consistency \citep{manakul2023selfcheckgptzeroresourceblackboxhallucination}.
In autoformalization, recent work checks a formalization for semantic consistency with its source text.
\citet{li2024autoformalize} generate several candidate formalizations of a mathematical statement 
and select one by self-consistency, combining agreement among the candidates with the similarity 
of each re-informalized candidate to the original.
\citet{chen2026reform} train a model to critique a formalization directly against the source 
text and to self-correct over successive revisions.
\citet{lu2025formalalign} train a scorer for the same purpose, which yields a learned 
alignment estimate rather than a decision.
In each case the judgment applies to a formalization as a whole.
We instead localize the failure to one translation step and repair only that step (\S\ref{sec:repair}).

\section{Roundtrip Autoformalization Framework}
\label{sec:framework}

\paragraph{Target formalism and notation.}
Let $\mathcal{X}$ denote the set of natural-language strings.
We use $\Sigma$ to represent the target formalism.
The framework is agnostic to the specific choice of $\Sigma$: all that is
required is for there to be a notion of well-formed expressions and a notion of
formal equivalence.  Let $\mathcal{Y}_{\Sigma}$ denote the set of well-formed $\Sigma-$expressions.
We write $y_a \equiv_{\Sigma} y_b$ when two formulas $y_a, y_b \in
\mathcal{Y}_{\Sigma}$ are equivalent according to $\Sigma$.
For example, $\Sigma$ could be first-order logic with the
standard notion of logical equivalence, or it could be a formalism based on
dependent type theory such as that used by the Lean~4 theorem prover~\citep{moura2021lean4}.

\paragraph{Three translation stages.}
We model roundtrip autoformalization as a composition of three translation functions:
\begin{align}
T_1 &: \mathcal{X} \rightarrow \mathcal{Y}_{\Sigma} \\
T_2 &: \mathcal{Y}_{\Sigma} \rightarrow \mathcal{X} \\
T_3 &: \mathcal{X} \rightarrow \mathcal{Y}_{\Sigma}
\end{align}
where $T_1$ performs \emph{autoformalization}, $T_2$ performs \emph{back-translation}, and $T_3$ performs \emph{re-formalization}. 
All three stages operate under the same fixed target formalism $\Sigma$.
Each $T_i$ is assumed to be \emph{stateless}: successive calls are mutually independent, with no shared memory or parameter updates between stages.
The framework is agnostic to how each $T_i$ is realized (e.g., an LLM, a rule-based translator, or a human annotator).
Note that $T_1$ and $T_3$ share the same type: both translate from natural
language strings to well-formed $\Sigma$ expressions. We consider them distinct
to maintain generality and also so that stage-level diagnosis and scoped repair can refer unambiguously to the pipeline stage at which an error originates.

\paragraph{Roundtrip instance.}
Consider again Figure~\ref{fig:roundtrip}.
Given an input $x \in \mathcal{X}$, the roundtrip pipeline produces:
\begin{align}
y^{\mathrm{orig}} &= T_1(x) \in \mathcal{Y}_{\Sigma} \\
x' &= T_2(y^{\mathrm{orig}}) \in \mathcal{X} \\
y^{\mathrm{rt}} &= T_3(x') \in \mathcal{Y}_{\Sigma}.
\end{align}
We refer to $y^{\mathrm{orig}}$ as the \emph{original formalization} and to $y^{\mathrm{rt}}$ as the \emph{roundtrip formalization}.

\paragraph{Why roundtrip?}
The roundtrip construction yields a verification signal \emph{without requiring ground-truth formalizations}.
If $T_1$ faithfully captures the meaning of $x$, then back-translating and re-formalizing should recover the same formal semantics.
When the two formalizations disagree, the point of divergence localizes the error.


\paragraph{Formal equivalence.}
The primary verification question is whether $y^{\mathrm{orig}} \equiv_{\Sigma} y^{\mathrm{rt}}$.
If not, at least one translation stage has introduced or lost meaning, triggering a diagnosis step in which an LLM judge examines all four pipeline artifacts ($x, y^{\mathrm{orig}}, x', y^{\mathrm{rt}}$) to identify the responsible stage and produce a scoped explanation for repair (\S\ref{sec:repair}).
Note that if the roundtrip is consistent with the original (i.e., $y^{\mathrm{orig}} \equiv_{\Sigma} y^{\mathrm{rt}}$), correctness is still not guaranteed: the pipeline may stabilize at a semantically different fixed point where both formalizations agree yet neither faithfully represents~$x$.
We therefore treat formal consistency as \emph{necessary but not sufficient}.
In \S\ref{sec:nli}, we discuss how to supplement roundtrip verification with an additional NLI-based check.

\paragraph{Syntactic well-formedness as prerequisite.}
Before semantic equivalence can be checked, each formula must be syntactically
valid according to $\Sigma$.
LLM-generated formulas occasionally violate grammar or type constraints. We
allow up to five LLM-based correction attempts per formula, which in our
experience suffices to ensure that at least one parsable encoding is produced.

\section{Stage-Based Diagnosis and Iterative Repair}
\label{sec:repair}
When $y^{\mathrm{orig}} \not\equiv_{\Sigma} y^{\mathrm{rt}}$, the roundtrip instance is formally inconsistent, implying that at least one of the translation stages introduced semantic drift.
However, the equivalence test alone does not identify which stage is responsible.
We therefore introduce a stage-based diagnosis and repair procedure that localizes the failure and applies a targeted repair operator (Figure~\ref{fig:framework}).

\subsection{First-Failure Diagnosis}
\label{subsec:diagnosis}

\paragraph{First-failed stage.}
We view the pipeline as an ordered sequence of stages $T_1 \prec T_2 \prec T_3$.
Intuitively, an error introduced earlier may propagate downstream, so we aim to identify the earliest stage at which the pipeline ceases to preserve semantics.
Formally specifying natural-language semantics is beyond the scope of this work. Instead, diagnosis operates over the observable artifacts $(x, y^{\mathrm{orig}}, x', y^{\mathrm{rt}})$ and returns an index in $\{1,2,3\}$.

\paragraph{Diagnosis function.}
Let
\begin{equation}
D:\; \mathcal{X} \times \mathcal{Y}_{\Sigma} \times \mathcal{X} \times \mathcal{Y}_{\Sigma} \rightarrow \{1,2,3\} \times \mathcal{E}
\end{equation}
be a diagnosis procedure which, given the roundtrip artifacts, predicts the first failed stage and produces an
\emph{explanation} $e \in \mathcal{E}$, where $\mathcal{E}$ is the set of
possible explanations.
In our system, $D$ is implemented as a constrained LLM-based judge and
$\mathcal{E}$ is simply $\mathcal{X}$, the set of natural language strings.
We enforce a \emph{sequential checking protocol}: $D$ must examine stages strictly in order (first comparing $x$ with $y^{\mathrm{orig}}$ for Stage~1, then $y^{\mathrm{orig}}$ with $x'$ for Stage~2, and finally $x'$ with $y^{\mathrm{rt}}$ for Stage~3), halting at the first detected inconsistency.

\paragraph{Scoped diagnostic reasoning.}
Critically, the explanation $e$ returned by $D$ is constrained to reference \emph{only} the artifacts relevant to the diagnosed stage.
For instance, if Stage~2 is diagnosed as faulty, the explanation must describe the mismatch between $y^{\mathrm{orig}}$ and $x'$ without mentioning $x$ or $y^{\mathrm{rt}}$.
This scoping ensures that the diagnostic feedback can be directly incorporated into the corresponding repair prompt without introducing confounding information from other pipeline stages.

\subsection{Targeted Repair Operators}
\label{subsec:operators}

\paragraph{Repair operators.}
A key design principle is that repairing stage~$i$ invalidates all downstream
artifacts produced by later stages; after repair, we therefore regenerate all subsequent stages to maintain a coherent pipeline state.
Each repair operator receives the relevant artifacts together with the diagnostic explanation $e$ produced by $D$:
\begin{align}
R_1 &: \mathcal{X} \times \mathcal{Y}_{\Sigma} \times \mathcal{E} \rightarrow \mathcal{Y}_{\Sigma} \\
R_2 &: \mathcal{Y}_{\Sigma} \times \mathcal{X} \times \mathcal{E} \rightarrow \mathcal{X} \\
R_3 &: \mathcal{X} \times \mathcal{Y}_{\Sigma} \times \mathcal{E} \rightarrow \mathcal{Y}_{\Sigma}
\end{align}
where $R_1$ repairs the original formalization, $R_2$ repairs the reconstructed NL, and $R_3$ repairs the roundtrip formalization.
Each $R_i$ is instantiated as an LLM call operating under the same target
formalism $\Sigma$.
The diagnostic explanation $e$ is injected into the repair prompt, instructing the model to address the \emph{specific} issues identified during diagnosis rather than regenerating blindly.
This feedback-driven design enables targeted corrections: the repair prompt explicitly states what semantic mismatch was detected, and the LLM is required to resolve that mismatch while preserving other aspects of the encoding.

\begin{figure*}[t]
\centering
\begin{tikzpicture}[
    >=Stealth,
    nlbox/.style={draw, rounded corners=3pt, fill=blue!10, minimum width=1.5cm, minimum height=0.65cm, align=center, font=\small, inner sep=2pt},
    fmlbox/.style={draw, rounded corners=3pt, fill=orange!18, minimum width=1.5cm, minimum height=0.65cm, align=center, font=\small, inner sep=2pt},
    smtbox/.style={draw, rounded corners=3pt, fill=gray!12, minimum width=3.0cm, minimum height=0.95cm, align=center, font=\small, inner sep=3pt},
    succbox/.style={draw, rounded corners=3pt, fill=green!18, minimum width=3.0cm, minimum height=0.65cm, align=center, font=\small\bfseries, inner sep=3pt},
    diagbox/.style={draw, rounded corners=3pt, fill=red!12, minimum width=5.5cm, minimum height=0.85cm, align=center, font=\small, inner sep=4pt},
    repbox/.style={draw, rounded corners=3pt, fill=green!14, minimum width=1.6cm, minimum height=1.0cm, align=center, font=\scriptsize, inner sep=3pt},
    paneltitle/.style={font=\small\bfseries, text=black!75, anchor=west},
    rolelabel/.style={font=\scriptsize, text=black!60, inner sep=1pt},
    panelframe/.style={draw=black!75, line width=0.5pt, rounded corners=4pt, inner sep=10pt},
    arrow/.style={->, semithick, black},
    darrow/.style={->, semithick, dashed, gray!70},
    satarrow/.style={->, thick, red!70!black},
    reparrow/.style={->, semithick, dashed, green!50!black},
    buslabel/.style={font=\scriptsize, fill=white, inner sep=2pt},
    branchlbl/.style={font=\scriptsize\itshape, text=red!60!black, fill=white, inner sep=1pt},
]

\node[nlbox]  at (0,    0) (x)    {$x$};
\node[fmlbox] at (1.85, 0) (yorig){$y^{\mathrm{orig}}$};
\node[nlbox]  at (3.7,  0) (xp)   {$x'$};
\node[fmlbox] at (5.55, 0) (yrt)  {$y^{\mathrm{rt}}$};

\node[rolelabel, below=1pt of x]     (xrole)     {NL input};
\node[rolelabel, below=1pt of yorig] (yorigrole) {Formal};
\node[rolelabel, below=1pt of xp]    (xprole)    {NL recon};
\node[rolelabel, below=1pt of yrt]   (yrtrole)   {Formal};

\draw[arrow] (x)     -- node[above, font=\small] {$T_1$} (yorig);
\draw[arrow] (yorig) -- node[above, font=\small] {$T_2$} (xp);
\draw[arrow] (xp)    -- node[above, font=\small] {$T_3$} (yrt);

\node[paneltitle] (titleA) at (-0.6, 1.55) {A.~Roundtrip autoformalization};

\node[smtbox] at (2.775, -2.2) (smt)
  {SMT solver\\[-1pt]{\small $y^{\mathrm{orig}} \equiv_\Sigma y^{\mathrm{rt}}$\,?}};

\draw[darrow] (yorig.south) -- (smt.north -| yorig.south);
\draw[darrow] (yrt.south) -- (5.55, -1.225) -- (3.575, -1.225) -- (3.575, -1.725);

\node[succbox] at (2.775, -3.6) (success) {\checkmark~Success: self-consistent};
\draw[arrow] (smt) -- node[right, font=\scriptsize] {UNSAT} (success);

\node[diagbox] at (10.0, -0.7) (D)
  {\textbf{Diagnosis $D$} (LLM judge)};

\node[paneltitle] (titleB) at (7.0, 1.55) {B.~Diagnosis and scoped repair};

\draw[satarrow] (smt.east) -- node[above, font=\scriptsize, text=red!70!black] {SAT} ++(2.0, 0) |- (D.west);

\draw[darrow, -] (x.north)     -- (x.north     |- 0,0.95);
\draw[darrow, -] (yorig.north) -- (yorig.north |- 0,0.95);
\draw[darrow, -] (xp.north)    -- (xp.north    |- 0,0.95);
\draw[darrow, -] (yrt.north)   -- (yrt.north   |- 0,0.95);
\draw[darrow, -] (x.north |- 0,0.95) -- (yrt.north |- 0,0.95);
\draw[darrow] (yrt.north |- 0,0.95) -- node[buslabel, above, pos=0.5] {$x,\,y^{\mathrm{orig}},\,x',\,y^{\mathrm{rt}}$} (10, 0.95) -- (10, -0.5);

\node[repbox] at (7.7, -2.4) (r1)
  {$R_1$\\[-1pt]\scriptsize $x,\,y^{\mathrm{orig}},\,e$\\\scriptsize $\to$ new $y^{\mathrm{orig}}$};
\node[repbox] at (10.0, -2.4) (r2)
  {$R_2$\\[-1pt]\scriptsize $y^{\mathrm{orig}},\,x',\,e$\\\scriptsize $\to$ new $x'$};
\node[repbox] at (12.3, -2.4) (r3)
  {$R_3$\\[-1pt]\scriptsize $x',\,y^{\mathrm{rt}},\,e$\\\scriptsize $\to$ new $y^{\mathrm{rt}}$};

\draw[arrow] (D.south -| r1) -- node[branchlbl, pos=0.55] {if $i{=}1$} (r1.north);
\draw[arrow] (D.south -| r2) -- node[branchlbl, pos=0.55] {if $i{=}2$} (r2.north);
\draw[arrow] (D.south -| r3) -- node[branchlbl, pos=0.55] {if $i{=}3$} (r3.north);

\draw[reparrow, -] (r1.south) -- (r1.south |- 0,-5.0);
\draw[reparrow, -] (r2.south) -- (r2.south |- 0,-5.0);
\draw[reparrow, -] (r3.south) -- (r3.south |- 0,-5.0);
\draw[reparrow, -] (r1.south |- 0,-5.0) -- (r3.south |- 0,-5.0);
\draw[reparrow] (r1.south |- 0,-5.0) -- node[buslabel, above, text=green!40!black, pos=0.5] {rerun downstream $T_i$, and then SMT check} (-1.3, -5.0) -- (-1.3, 0) -- (-1, 0);

\coordinate (Abot) at (5.5, -4.05);
\coordinate (Bbot) at (10.0, -4.05);

\begin{scope}[on background layer]
  \node[panelframe, fit={(titleA) (x) (yrt) (yrtrole) (smt) (success) (Abot)}] (panelAframe) {};
  \node[panelframe, fit={(titleB) (D) (r1) (r3) (Bbot)}] (panelBframe) {};
\end{scope}

\end{tikzpicture}
\caption{The roundtrip autoformalization framework.
\textbf{(A)}~The pipeline produces two formal encodings of $x$ and an SMT solver checks them for equivalence.
\textbf{(B)}~On disagreement, diagnosis $D$ localizes the failed stage and repair operator $R_i$ corrects it before re-checking.}
\label{fig:framework}
\end{figure*}

\subsection{Iterative Repair Loop}
\label{subsec:loop}

\paragraph{Iterative procedure.}
We combine verification, diagnosis, and repair into an iterative loop that attempts to reach formal self-consistency within a bounded budget.
Starting from $(x, y^{\mathrm{orig}}, x', y^{\mathrm{rt}})$, we repeatedly:
(i) check whether $y^{\mathrm{orig}} \equiv_{\Sigma} y^{\mathrm{rt}}$,
(ii) if not, diagnose the first-failed stage, apply the corresponding repair operator, and regenerate downstream artifacts.
The loop terminates when equivalence is achieved (success) or after a fixed maximum number of iterations (failure).
We reiterate that when the procedure reports success, this does not guarantee correctness with respect to
the original natural-language intent, but it does provide a measure of reassurance and confidence.


\begin{algorithm}[t]
\caption{Roundtrip Verification and Stage-Based Repair}
\label{alg:roundtrip-repair}
\begin{algorithmic}[1]
\REQUIRE Natural-language input $x \in \mathcal{X}$, formalism $\Sigma$, max iterations $K$
\STATE $y^{\mathrm{orig}} \leftarrow T_1(x)$
\STATE $x' \leftarrow T_2(y^{\mathrm{orig}})$
\STATE $y^{\mathrm{rt}} \leftarrow T_3(x')$
\FOR{$k = 1$ to $K$}
    \IF{$y^{\mathrm{orig}} \equiv_{\Sigma} y^{\mathrm{rt}}$}
        \STATE \textbf{Return SUCCESS}, $(y^{\mathrm{orig}}, x', y^{\mathrm{rt}})$
    \ENDIF
    \STATE $i \leftarrow D(x, y^{\mathrm{orig}}, x', y^{\mathrm{rt}})$ \COMMENT{first-failed stage}
    \IF{$i = 1$}
        \STATE $y^{\mathrm{orig}} \leftarrow R_1(x, y^{\mathrm{orig}})$
        \STATE $x' \leftarrow T_2(y^{\mathrm{orig}})$
        \STATE $y^{\mathrm{rt}} \leftarrow T_3(x')$
    \ELSIF{$i = 2$}
        \STATE $x' \leftarrow R_2(y^{\mathrm{orig}}, x')$
        \STATE $y^{\mathrm{rt}} \leftarrow T_3(x')$
    \ELSIF{$i = 3$}
        \STATE $y^{\mathrm{rt}} \leftarrow R_3(x', y^{\mathrm{rt}})$
    \ENDIF
\ENDFOR
\STATE \textbf{Return FAILURE}, $(y^{\mathrm{orig}}, x', y^{\mathrm{rt}})$
\end{algorithmic}
\end{algorithm}

\section{Auxiliary Semantic Signal}
\label{sec:nli}

\paragraph{Bidirectional NLI as a semantic proxy.}
We use natural language inference (NLI) to compare the original rule $x$ with the reconstructed natural-language description $x' = T_2(y^{\mathrm{orig}})$.
NLI has been used as a faithfulness proxy across NLG tasks \citep{falke-etal-2019-ranking, honovich-etal-2022-true-evaluating, laban-etal-2022-summac}.
If the roundtrip preserves semantics, $x$ and $x'$ should be mutually entailing: $x$ entails $x'$ (nothing lost) and $x'$ entails $x$ (nothing spuriously added).
We operationalize this using a BART-large model fine-tuned on MultiNLI, computing entailment probabilities in both directions:
\begin{align}
E_{\rightarrow} &= P(\text{entailment} \mid x, x') \\
E_{\leftarrow} &= P(\text{entailment} \mid x', x)
\end{align}
along with contradiction probabilities $C_{\rightarrow}$ and $C_{\leftarrow}$.

\paragraph{Diagnostic categories.}
We derive $E_{\min} = \min(E_{\rightarrow}, E_{\leftarrow})$ for mutual entailment and $C_{\max} = \max(C_{\rightarrow}, C_{\leftarrow})$ for contradiction.
Each rule pair is classified into one of six categories via the decision tree in Figure~\ref{fig:nli-tree} (first match wins).
The six categories (\emph{Contradiction}, \emph{Equivalent}, \emph{Strengthened}, \emph{Weakened}, \emph{Related}, \emph{Unrelated}) capture not only whether meaning is preserved but \emph{how} it drifts, a distinction that a binary test would obscure.

\begin{figure}[t]
\centering
\fcolorbox{gray!40}{white}{%
\begin{tikzpicture}[
  test/.style={draw, rounded corners=2pt, fill=gray!8,
               font=\footnotesize, inner sep=3.5pt,
               minimum height=0.42cm, anchor=west},
  cat/.style={font=\footnotesize\bfseries, anchor=west,
              fill=white, inner sep=2pt},
  fall/.style={font=\scriptsize, anchor=east, text=black!60},
  >=Stealth,
  node distance=0.38cm,
]
\node[test] (c1) {$C_{\max} \!\geq\! 0.6$};
\node[test, below=of c1] (c2)
  {$E_{\min} \!\geq\! 0.7\;\wedge\; C_{\max} \!<\! 0.2$};
\node[test, below=of c2] (c3)
  {$E_{\!\rightarrow} \!\geq\! 0.6\;\wedge\; E_{\!\leftarrow} \!<\! 0.4$};
\node[test, below=of c3] (c4)
  {$E_{\!\leftarrow} \!\geq\! 0.6\;\wedge\; E_{\!\rightarrow} \!<\! 0.4$};
\node[test, below=of c4] (c5)
  {$E_{\min} \!\geq\! 0.3\;\wedge\; C_{\max} \!<\! 0.3$};

\node[cat, right=0.55cm of c1.east] (l1) {Contradiction};
\node[cat, right=0.55cm of c2.east] (l2) {Equivalent};
\node[cat, right=0.55cm of c3.east] (l3) {Strengthened};
\node[cat, right=0.55cm of c4.east] (l4) {Weakened};
\node[cat, right=0.55cm of c5.east] (l5) {Related};

\node[cat, below=0.32cm of c5.south, anchor=north] (l6) {Unrelated};

\foreach \i/\lab in {1/l1, 2/l2, 3/l3, 4/l4, 5/l5} {
  \draw[->, thick, black!70]
    (c\i.east) -- node[above=-1pt, font=\scriptsize, text=black!50] {yes} (\lab);
}

\foreach \i/\j in {1/c2, 2/c3, 3/c4, 4/c5} {
  \draw[->, thick, black!70]
    (c\i.south) -- node[fall] {no\;\;} (\j.north);
}
\draw[->, thick, black!70]
    (c5.south) -- node[fall] {no\;\;} (l6.north);
\end{tikzpicture}%
}
\caption{NLI diagnostic decision tree (first match wins).
Thresholds are conservative. \emph{Equivalent} demands strong bidirectional entailment (${\geq}\,0.7$) with near-zero contradiction (${<}\,0.2$), while \emph{Contradiction} requires high conflict (${\geq}\,0.6$).}
\label{fig:nli-tree}
\end{figure}

\paragraph{Diagnostic, not supervisory.}
NLI scores are used \emph{only for post-hoc analysis}, not as optimization targets.
The repair loop (\S\ref{sec:repair}) operates solely on formal equivalence. NLI serves as an independent diagnostic lens, allowing us to ask: \emph{when formal self-consistency is achieved, does semantic fidelity follow?}
This separation avoids optimizing for a noisy proxy while still providing a meaningful signal.
In \S\ref{sec:results}, we cross-tabulate formal equivalence with NLI categories to reveal where formal consistency diverges from semantic alignment.

\section{Experimental Setup}
\label{sec:experiments}

\subsection{Domains and Datasets}

\paragraph{From law to logic.}
We instantiate the framework on statutory regulation, which combines natural-language ambiguity with precise operational requirements.
Applying statutes to facts is an established NLP task \citep{holzenberger2020datasetstatutoryreasoningtax}.
LLMs reason over statutes unreliably \citep{10.1145/3594536.3595163}, and their formalizations of legal text introduce axioms absent from the source \citep{wang2026knowlimitsfaithfulness}.
We evaluate on two corpora: traffic law and wildlife regulation.
Any statutory domain admitting a formal equivalence check could serve equally well (\S\ref{sec:framework}).

\paragraph{Traffic corpus.}
We evaluate on 150 rules drawn from the Texas Transportation Code.
These rules govern vehicle behavior on roadways, covering lane positioning, passing maneuvers, signaling requirements, speed limits, and interactions with special vehicles (e.g., streetcars, school buses, emergency vehicles).

\paragraph{Wildlife corpus.}
We evaluate on 77 rules drawn from the Texas Parks and Wildlife Code, a second statutory domain.
These rules cover hunting and fishing licensing, season and bag-limit restrictions, equipment and method restrictions, and protected-species rules.

\subsection{Models and Configuration}

\paragraph{Formalism.}
A \emph{Satisfiability Modulo Theories} (SMT) solver is a tool that can
formally determine whether a set of first-order formulas is \emph{satisfiable}
(i.e., whether there exists an interpretation that makes all of the formulas
true) with respect to some background theory $T$~\citep{barrett2010smtlib}.
An SMT solver returns one of two verdicts: \textsc{sat} (a satisfying interpretation exists) or \textsc{unsat} (no such interpretation exists).
SMT solvers can check whether two formulas $\varphi$ and $\psi$ are logically
equivalent (with respect to a theory $T$), 
by checking the satisfiability of $\lnot(\varphi = \psi)$. If the result is
\textsc{unsat}, no interpretation can distinguish the two formulas, certifying equivalence. 
If \textsc{sat}, the solver can return a counterexample demonstrating the difference.
We construct a \emph{domain schema} for each corpus, a set of SMT declarations and definitions that provide a fixed vocabulary for that domain's concepts.
Schemas are produced semi-automatically: we draft a sketch by hand, scale it up with an LLM, and curate the result against a list of design principles to avoid common issues (Appendix~\ref{sec:schema}).
We use the Z3 SMT solver~\citep{demoura2008z3} (version 4.15.3) to check equivalence, and all queries resolve in under one second.

\paragraph{Language models and repair.}
To assess framework generalizability across model families, we run the full pipeline with two frontier LLMs:
\textbf{Claude Opus~4.6} (Anthropic) and \textbf{GPT-5.2} (OpenAI).
Both use temperature 0.3.
Each translation stage ($T_1$, $T_2$, $T_3$) and repair operator ($R_1$, $R_2$, $R_3$) is realized as a single stateless LLM call. The domain schema is injected into every prompt, and no context is carried between calls.
All stages and operators use the same prompts for both models.
The iterative repair loop (\S\ref{sec:repair}) runs for at most $K{=}3$ iterations per rule, since gains flatten quickly.
Each reported result is from a single run.
We re-ran every experiment twice more, end to end, and report the spread in Appendix~\ref{sec:variance}.

\paragraph{NLI model.}
For semantic comparison between original and reconstructed natural language, we use BART-large fine-tuned on MultiNLI \citep{lewis2020bart,williams2018mnli} (\texttt{facebook/bart-large-mnli}).
Inference is performed bidirectionally as described in \S\ref{sec:nli}.

\paragraph{Compute.}
Claude Opus~4.6 and GPT-5.2 are commercial APIs whose parameter counts are not publicly disclosed.
NLI inference uses BART-large (approximately 0.4B parameters) on a single laptop.
Per-repair LLM call counts, the dominant cost, are reported in Table~\ref{tab:repair-summary}.

\subsection{Ablation Conditions}
\label{subsec:ablation}

We compare four approaches:
\begin{enumerate}
\item \textbf{No Repair} (Baseline~0): The roundtrip pipeline runs without any repair loop. The initial equivalence check result is final.
\item \textbf{Random Stage} (Baseline~1): When the equivalence check fails, select a stage $i \in \{1,2,3\}$ uniformly at random, repair it, and regenerate downstream stages. Repeat up to $K{=}3$ times, stopping as soon as equivalence holds. No diagnosis is performed.
\item \textbf{Regenerate} (Baseline~2): When the equivalence check fails, regenerate the entire pipeline from the original input $x$ and check again. Repeat up to $K{=}3$ times, stopping as soon as equivalence holds. No diagnosis is performed.
\item \textbf{Full} (Ours): We use the complete pipeline with diagnosis (\S\ref{subsec:diagnosis}) followed by targeted repair of the diagnosed stage. This is the system described in Algorithm~\ref{alg:roundtrip-repair}.
\end{enumerate}
All four approaches share the same initial roundtrip pass (identical $y^{\mathrm{orig}}, x', y^{\mathrm{rt}}$). They differ only in the repair strategy applied to rules where the initial equivalence check returns SAT.

The two new baselines isolate two design choices in Full. Random Stage controls for the value of \emph{diagnosis} (which stage is selected for repair). Regenerate controls for the value of \emph{scoping} (re-running only the diagnosed stage rather than the whole pipeline).

\section{Results and Analysis}
\label{sec:results}

\subsection{Roundtrip Equivalence and Repair}

Table~\ref{tab:repair-summary} reports final equivalence outcomes for each experiment.
Without repair, equivalence ranges from 44.7\,\% (Traffic / Claude) to 67.1\,\% (Wildlife / GPT).
Each repair method substantially raises equivalence.

\paragraph{Method comparison.}
The most effective repair method depends on the model, and the same pattern holds across both domains.
For Claude, Full leads on both UNSAT count and cost-per-repair.
For GPT with same-model diagnosis, Regenerate surpasses Full on UNSAT count, and Full is the most expensive of the 
three methods.
This split reflects the diagnosis distribution: GPT repeatedly diagnoses $T_1$, which forces regenerating 
$T_2$ and $T_3$ downstream on every iteration (\S\ref{subsec:stage}).
The third row of each GPT block reports results from a cross-model approach, where Claude performs only the diagnosis 
step. We analyze this approach in \S\ref{subsec:gpt-collapse}.

\begin{table*}[t]
\centering
\footnotesize
\setlength{\tabcolsep}{4pt}
\begin{tabular}{@{}lllcccc@{\hskip 8pt}ccc@{}}
\toprule
& & & \multicolumn{4}{c}{\textbf{Final UNSAT count (\%)}} & \multicolumn{3}{c}{\textbf{LLM calls / repair}} \\
\cmidrule(lr){4-7} \cmidrule(l){8-10}
\textbf{Domain} & \textbf{Model} & \textbf{Diagnoser} & \textbf{None} & \textbf{Random} & \textbf{Regen} & \textbf{Full} & \textbf{Random} & \textbf{Regen} & \textbf{Full} \\
\midrule
\multirow{3}{*}{Traffic ($N{=}150$)}
 & Claude & Claude & 67~(44.7) & 100~(66.7) & 97~(64.7) & \textbf{128~(85.3)} & 12.27 & 19.90 & \textbf{7.21} \\
 & GPT    & GPT    & 92~(61.3) & 118~(78.7) & \textbf{127~(84.7)} & 124~(82.7) & \textbf{9.77} & 10.54 & 14.56 \\
 & GPT    & Claude & --- & --- & --- & \textbf{136~(90.7)} & --- & --- & \textbf{7.86} \\
\midrule
\multirow{3}{*}{Wildlife ($N{=}77$)\textsuperscript{$\dagger$}}
 & Claude & Claude & 48~(62.3) & 60~(77.9) & 63~(81.8) & \textbf{66~(85.7)} & 11.67 & 12.20 & \textbf{9.50} \\
 & GPT    & GPT    & 51~(67.1) & 59~(77.6) & \textbf{65~(85.5)} & 60~(78.9) & 12.75 & \textbf{11.57} & 26.44 \\
 & GPT    & Claude & --- & --- & --- & \textbf{72~(94.7)} & --- & --- & \textbf{6.05} \\
\bottomrule
\end{tabular}
\caption{Roundtrip equivalence outcomes across two domains and four repair conditions. The \textbf{Model} column 
performs pipeline and repair. The \textbf{Diagnoser} performs diagnosis. Bold marks the per-row best. 
\textsuperscript{$\dagger$}Claude rows use all 77 rules. GPT rows use $N{=}76$: one rule yielded no Stage-1 encoding under GPT and is excluded from every GPT count.}
\label{tab:repair-summary}
\end{table*}

\subsection{Stage Diagnosis Distribution}
\label{subsec:stage}

The two models diagnose pipeline failures very differently, and the pattern is consistent across both 
domains (Table~\ref{tab:stage-diag}, Appendix~\ref{sec:stage-appendix}).
Claude distributes its diagnoses across all three stages.
GPT collapses almost all of its diagnoses onto $T_1$.
The Random baseline distributes approximately uniformly across stages in every case, so the 
concentration is a property of the diagnosis function, not the data.
The distribution does not say whether those diagnoses are correct.
Appendix~\ref{sec:human-eval} reports a human evaluation of the diagnosed stage.

Repairing $T_1$ cascades: it forces a fresh $T_2$ and a fresh $T_3$ on every iteration, while 
repairing $T_3$ does not.
GPT's $T_1$-heavy diagnosis therefore behaves like full-pipeline regeneration plus a diagnosis 
call, which explains Full's higher cost than Regenerate for GPT (Table~\ref{tab:repair-summary}).
\S\ref{subsec:gpt-collapse} tests whether this $T_1$ concentration is a property of the protocol or of the model.

\subsection{Formal vs.\ Semantic Alignment}
\label{subsec:nli-results}

\begin{figure*}[t]
  \centering
  \includegraphics[width=0.72\textwidth]{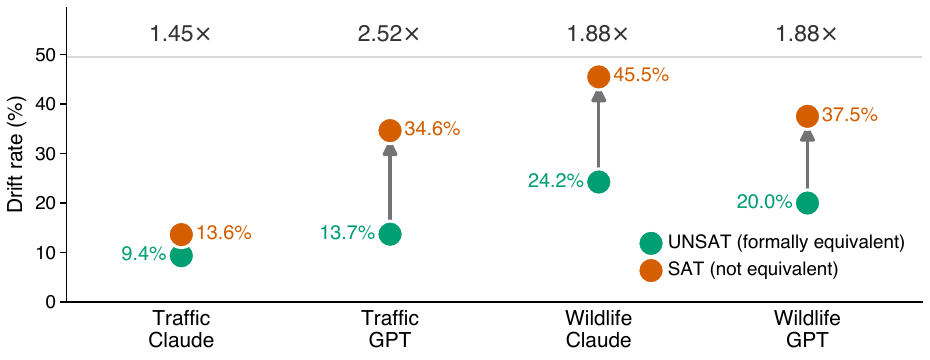}
  \caption{NLI drift rate among UNSAT (formally equivalent) and SAT (not equivalent) post-repair rules
  under Full repair, one domain-model pair.
  The annotation above each pair is the SAT-to-UNSAT drift ratio.}
  \label{fig:drift-bars}
\end{figure*}

\begin{table}[t]
\centering
\scriptsize
\setlength{\tabcolsep}{3pt}
\begin{tabular}{@{}llccc@{}}
\toprule
\textbf{Domain / Model} & \textbf{Method} & \textbf{UNSAT} & \textbf{SAT} & \textbf{Ratio} \\
\midrule
\multirow{3}{*}{Traffic / Claude}
 & Random      & 14.0\% & 20.0\% & 1.43 \\
 & Regenerate  & 12.4\% & 20.8\% & 1.68 \\
 & Full        & \phantom{0}9.4\% & 13.6\% & 1.45 \\
\midrule
\multirow{4}{*}{Traffic / GPT}
 & Random              & 15.3\% & 28.1\% & 1.84 \\
 & Regenerate          & 15.0\% & 26.1\% & 1.74 \\
 & Full                & 13.7\% & 34.6\% & 2.52 \\
 & Full~(Claude~dx)\textsuperscript{$\ast$}    & 12.5\% & 35.7\% & 2.86 \\
\midrule
\multirow{3}{*}{Wildlife / Claude}
 & Random      & 25.0\% & 35.3\% & 1.41 \\
 & Regenerate  & 25.4\% & 28.6\% & 1.13 \\
 & Full        & 24.2\% & 45.5\% & 1.88 \\
\midrule
\multirow{4}{*}{Wildlife / GPT}
 & Random              & 18.6\% & 35.3\% & 1.89 \\
 & Regenerate          & 20.0\% & 54.5\% & 2.73 \\
 & Full                & 20.0\% & 37.5\% & 1.88 \\
 & Full~(Claude~dx)\textsuperscript{$\ast$}    & 19.4\% & 75.0\%\textsuperscript{$\ddagger$} & 3.86\textsuperscript{$\ddagger$} \\
\midrule
\textbf{Pooled} & & \textbf{16.2\%} & \textbf{28.7\%} & \textbf{1.78} \\
\bottomrule
\end{tabular}
\caption{Post-repair drift rate (share of rules NLI-classified as Unrelated or Contradiction), split by SMT verdict.
SAT denotes every rule not verified equivalent.
Ratio is SAT drift divided by UNSAT drift.
\textsuperscript{$\ast$}Cross-model: GPT pipeline + Claude diagnosis (\S\ref{subsec:gpt-collapse}).
\textsuperscript{$\ddagger$}This SAT pool has only 4 rules, so the rate is noisy.}
\label{tab:drift-summary}
\end{table}

We examined the rate of every NLI category in each pool (Appendix~\ref{sec:nli-appendix}, Table~\ref{tab:per-cat}).
Related and Weakened distribute similarly between UNSAT and SAT.
Strengthened concentrates in UNSAT.
This reflects an artifact of the back-translation: the reconstructed NL tends to use the schema's predicate 
names verbatim, which produces wordier text than the original lawyer-written rule.
NLI interprets the wordier reconstruction as a stricter version of the original, so NLI-Strengthened arises 
even when the formalization is faithful.
Only Unrelated and Contradiction discriminate cleanly between UNSAT and SAT, motivating their use as our Drift measure.

\paragraph{Formal equivalence predicts semantic faithfulness.}
Figure~\ref{fig:drift-bars} shows that under Full repair, SAT rules drift 1.45$\times$-2.52$\times$ more 
often than UNSAT rules across the four domain-model pairs.
Pooled across these four Full configurations, SAT rules drift 2.03 times more often than UNSAT rules.
The same pattern holds when baselines are included: pooled across all 14 rows in 
Table~\ref{tab:drift-summary}, SAT rules still drift 1.78 times more often than UNSAT rules.
The pattern is consistent across both model families and both statutory domains.
Appendix~\ref{sec:human-eval} reports a human evaluation of faithfulness against the equivalence verdict.
Appendix~\ref{sec:pre-repair} shows the same gap before any repair runs.
Formal equivalence does not guarantee semantic fidelity, but it predicts it.

\subsection{Residual Failure Analysis}
\label{subsec:failure}

A rule is \emph{residual} if none of Random, Regenerate, or Full repairs it within $K{=}3$ iterations. 26 unique rules are 
residual across the four domain-model pairs. Manual inspection identifies four recurring patterns of 
intrinsic rule difficulty: \textbf{schema vocabulary gaps} (the schema lacks a needed predicate, so $T_1$ and $T_3$ 
approximate with different substitutes), \textbf{compound conditions} (disjunctive or nested premises where $T_1$ and 
$T_3$ encode the AND/OR structure differently), \textbf{exception clauses} (rules that suspend an obligation under a 
condition, where $T_2$ back-translates the exception as a positive restatement and $T_3$ then encodes the inverted meaning), 
and \textbf{conditional thresholds} (numeric thresholds that depend on another variable, where the pipeline often 
drops the dependency). The bottleneck is intrinsic rule structure, not iteration budget. 

\subsection{Why does GPT's diagnosis collapse?}
\label{subsec:gpt-collapse}

The GPT/Full underperformance we report in \S\ref{subsec:stage} stems from a single design point: the 
diagnosis function. GPT selects the first translation step $T_1$ in 83\,\% of iterations on 
Traffic and 97\,\% on Wildlife, regardless of where the actual error occurs (Table~\ref{tab:stage-diag}). 
Two hypotheses are compatible with this: the prompt's sequential checking protocol biases GPT toward the 
first stage examined, or GPT has an intrinsic preference for blaming initial formalization. We test both directly.

\paragraph{Non-sequential prompt.}
We rewrite the diagnosis prompt to drop the ``examine $T_1$ first, then $T_2$, then $T_3$, halt at first 
failure'' instruction and ask the judge to consider all three stages together. Re-running diagnosis on 
the existing GPT SAT pools, $T_1$ selection drops from 84\,\% to 9\,\% on Traffic and from 96\,\% to
16\,\% on Wildlife. These baselines are first-iteration rates over the re-diagnosed pools of 58 and
25 rules, so they differ from the 83\,\% and 97\,\% pooled over all iterations in
Table~\ref{tab:stage-diag}. The protocol explains most of the collapse.

\paragraph{Cross-model diagnosis.}
We rerun the Full repair loop on the existing GPT pipeline outputs and change one thing:
Claude acts as the diagnosis judge, while GPT still writes every formalization and performs
every repair.
Equivalence rises on both domains and cost per repair falls on both
(Table~\ref{tab:repair-summary}, last row of each block).
Scoped repair recovers the lead it had lost to Regenerate.
The diagnosis function was the bottleneck, not the pipeline or the repair operators.

\section{Conclusion}
\label{sec:conclusion}

We evaluated a roundtrip verification and scoped repair framework on two statutory domains and two model families.
Two findings stand out.
First, formal equivalence is a useful predictor of semantic faithfulness: across the four Full configurations spanning both 
domains and both models, SAT rules show 1.4$\times$-2.5$\times$ more NLI drift than UNSAT rules.
Second, diagnosis-guided scoped repair achieves the highest verified-equivalence rate at the lowest cost when the 
diagnosis function is reliable.
On Claude pipelines, same-model scoped repair wins on both domains.
On GPT pipelines, the diagnosis step is the bottleneck for scoped repair (not the pipeline or repair operators): 
replacing only the diagnoser with a different model restores scoped repair's lead at lower cost (\S\ref{subsec:gpt-collapse}).
About 16\,\% of formally equivalent rules still drift on NLI.
Future work could integrate additional checks into the repair loop, such as NLI or an LLM-based check, rather than relying on formal equivalence alone.

\section*{Limitations}
\label{sec:limitations}

\paragraph{Domain scope.}
Both corpora are English statutory text from Texas.
Mathematical autoformalization benchmarks typically operate one theorem at a time, where a human author can read the formal statement and verify it against the informal one.
Statutory text involves a large domain schema (close to 200 predicates) and operational ambiguity, so hand verification at scale is impractical and a ground-truth-free faithfulness signal is most directly motivated by this setting.
We do not claim the framework transfers to other genres such as code or scientific text without further evaluation.

\paragraph{Verification is heuristic, not a proof.}
The roundtrip check detects when $T_1(x)$ and $T_3(T_2(T_1(x)))$ disagree, but it cannot detect errors that affect both stages in the same way.
For example, if $T_1$ silently drops a conjunct of $x$ and $T_3$ reproduces the same omission from the back-translation, the equivalence check passes despite both being wrong.
The NLI cross-check (\S\ref{sec:nli}) is a partial mitigation, but it relies on a general-purpose model with known lexical-overlap biases \citep{mccoy2019right} on legal text, so its drift signal is itself noisy.

\paragraph{Rules are formalized in isolation.}
Statutory text can be heavily cross-referenced.
A rule may incorporate definitions from another section, suspend obligations stated elsewhere, or impose conditions on rules outside its own clause.
Our pipeline formalizes one rule at a time and treats each rule as self-contained.
Faithful formalization of densely cross-referenced rules likely requires presenting the connected cluster of rules together as input, rather than a single rule node.
We leave this to future work.

\section*{Risks}
\label{sec:risks}
Autoformalization outputs generated by this system, including those that
pass roundtrip verification, should not be used in safety-critical applications,
including but not limited to autonomous systems, medical devices, or safety case
documentation, without independent review by a qualified formal methods expert.
Regarding environmental impact, the pipeline relies on API-based LLM inference and a locally-run SMT solver. We did not train or fine-tune any models, so the compute footprint is bounded by inference calls.

\section*{Acknowledgements}
This work was supported in part by the Stanford Center for Automated Reasoning (Centaur).

\section*{Use of AI Assistants}
Claude Code was carefully used to assist with code implementation, \LaTeX{} formatting, and paraphrasing 
during paper writing. All scientific content, experimental design, analysis, and claims are entirely the authors' own.

\bibliography{references}

\appendix

\section{Prompt Templates}
\label{sec:prompts}

The pipeline uses prompts for each pipeline component: the three translation stages $T_1$, $T_2$, $T_3$, the diagnosis function $D$, and the three stage-specific repair operators $R_1$, $R_2$, $R_3$.
All prompts share a common structure: a system role description, the domain schema injected verbatim, the relevant pipeline artifacts, and explicit output-format constraints.
Below is the diagnosis prompt in abridged form.

\begin{lstlisting}[basicstyle=\scriptsize\ttfamily, caption={Diagnosis prompt (abridged).}]
You are analyzing a 3-arrow translation
pipeline that converts natural language
into SMT-LIB encodings:
  ARROW 1: Original NL -> Phase 1 SMT
  ARROW 2: Phase 1 SMT -> Reconstructed NL
  ARROW 3: Reconstructed NL -> Phase 3 SMT

Z3 returned SAT, so Phase 1 and Phase 3
encodings differ. Identify the FIRST arrow
that introduced the divergence.

Procedure:
1. Check Arrow 1. If Phase 1 SMT does not
   match Original NL semantics, stop and
   report Arrow 1.
2. Else check Arrow 2. If Reconstructed NL
   does not describe Phase 1 SMT, stop and
   report Arrow 2.
3. Else report Arrow 3.

ARTIFACTS:
  Original NL:        {original_nl}
  Phase 1 SMT:        {phase1_smt}
  Reconstructed NL:   {reconstructed_nl}
  Phase 3 SMT:        {phase3_smt}

Respond in this format:
  FIRST_FAILED_ARROW: [1, 2, or 3]
  REASONING: [a short paragraph; reference
    only the artifacts adjacent to the
    diagnosed arrow]
\end{lstlisting}

The repair prompts inject the diagnostic reasoning as a \texttt{DIAGNOSTIC FEEDBACK} field and instruct the model to change only what is necessary to address the identified problem.

\section{Schemas}
\label{sec:schema}

\paragraph{Construction process.}
Each domain schema is produced semi-automatically.
We hand-draft an initial schema sketch on a small set of pilot rules, then ask an LLM to extend the sketch with the additional sorts, predicates, and enumerated types needed to encode the full corpus.
We iterate (proposing, reviewing, refining) until coverage is satisfactory.
At each round we curate the result against the design principles described below, removing or rewriting predicates that violate them.

\paragraph{Design principles.}
We curate each schema against several principles to keep predicates reusable across rules and to avoid common smells.
Each predicate should encode a single concept rather than a multi-clause sentence.
Action verbs and qualifiers compose separately rather than fusing into one predicate.
Anchor predicates such as kind, used-device, or behavior tags are designed to fire in many rules.
Discrimination among related entities uses one sort with a kind enumeration rather than multiple sorts.
Numeric thresholds are exposed as Real-valued functions rather than baked into predicate names.
Opaque named enumerations (for example, long lists of protected species) collapse into a single black-box predicate.
Time-varying predicates take an integer time argument in the last position; time-independent predicates omit it.

\paragraph{Excerpts.}
Short illustrative excerpts of both schemas are shown below.
\begin{lstlisting}[basicstyle=\scriptsize\ttfamily, caption={Traffic schema (excerpt).}]
; Sorts
(declare-sort Vehicle 0)
(declare-sort Roadway 0)

; Static properties
(declare-fun kind (Vehicle) VehicleKind)
(declare-fun roadway_kind (Roadway) RoadwayKind)

; Time-varying predicates (Int = time)
(declare-fun on_roadway (Vehicle Roadway Int) Bool)
(declare-fun speed (Vehicle Int) Real)
\end{lstlisting}

\begin{lstlisting}[basicstyle=\scriptsize\ttfamily, caption={Wildlife schema (excerpt).}]
; Sorts
(declare-sort Person 0)
(declare-sort Animal 0)
(declare-sort Device 0)

; Static properties
(declare-fun is_kind (Animal AnimalKind) Bool)
(declare-fun protected_by_code (Animal) Bool)

; Time-varying predicates (Int = time)
(declare-fun in_captivity (Animal Int) Bool)
(declare-fun hunts (Person Animal Int) Bool)
\end{lstlisting}

\section{Stage Diagnosis Details}
\label{sec:stage-appendix}

\begin{table}[t]
\centering
\footnotesize
\setlength{\tabcolsep}{3pt}
\begin{tabular}{@{}llccc@{}}
\toprule
\textbf{Domain} & \textbf{Model / Method} & $T_1$ & $T_2$ & $T_3$ \\
\midrule
\multirow{6}{*}{Traffic}
 & Claude / Full                       & 25\% & 21\% & \textbf{55\%} \\
 & Claude / Random                     & 34\% & 35\% & 32\% \\
 & GPT / Full (sequential)             & \textbf{83\%} & \phantom{0}3\% & 14\% \\
 & GPT / Random                        & 32\% & 30\% & 38\% \\
 & GPT / Full (non-sequential)         & \phantom{0}9\% & 17\% & 74\% \\
 & GPT / Full (Claude judge)           & 48\% & 12\% & 39\% \\
\midrule
\multirow{6}{*}{Wildlife}
 & Claude / Full                       & 32\% & \phantom{0}8\% & \textbf{60\%} \\
 & Claude / Random                     & 35\% & 27\% & 38\% \\
 & GPT / Full (sequential)             & \textbf{97\%} & \phantom{0}3\% & \phantom{0}0\% \\
 & GPT / Random                        & 30\% & 19\% & 51\% \\
 & GPT / Full (non-sequential)         & 16\% & 20\% & 64\% \\
 & GPT / Full (Claude judge)           & 45\% & 12\% & 43\% \\
\bottomrule
\end{tabular}
\caption{Stage selection distribution under the diagnosis function (Full) and the random baseline (Random).
On both domains, Claude's diagnosis distributes across stages with a $T_3$ concentration, and GPT's diagnosis collapses onto $T_1$ under the sequential checking protocol.
Random selection is approximately uniform.
Two interventions un-collapse the GPT distribution: rewriting the prompt to drop sequential ordering (\emph{non-sequential}, \S\ref{subsec:gpt-collapse}, Experiment A), or replacing GPT with Claude as the judge (\emph{Claude judge}, \S\ref{subsec:gpt-collapse}, Experiment C).}
\label{tab:stage-diag}
\end{table}

\begin{figure*}[t]
  \centering
  \includegraphics[width=0.92\textwidth]{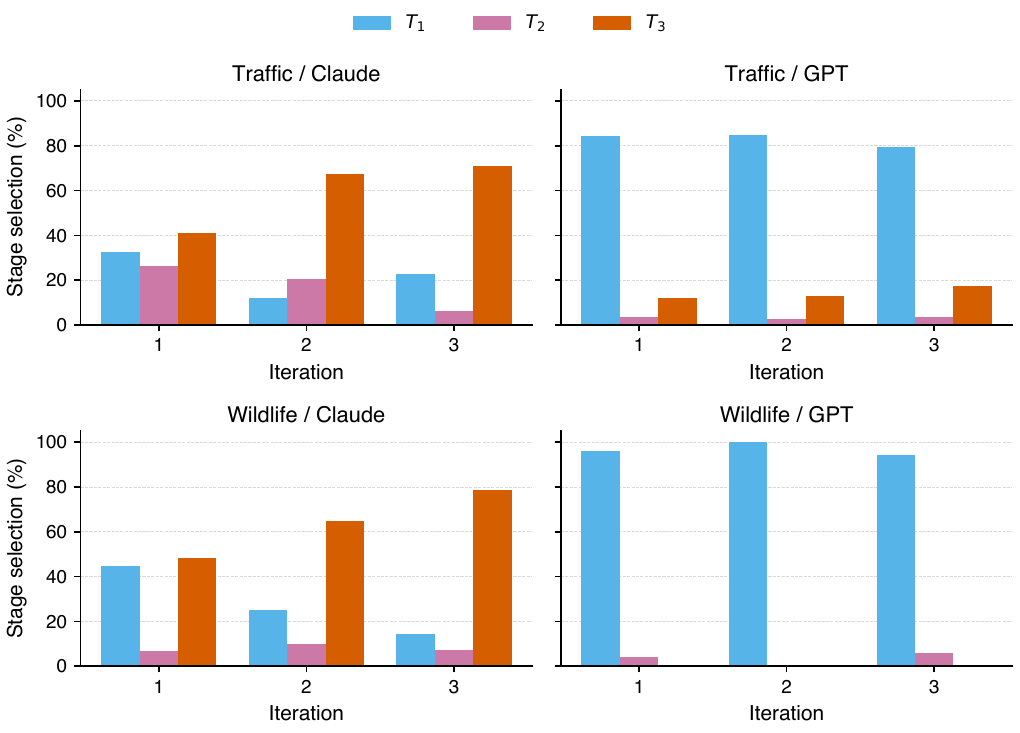}
  \caption{Per-iteration stage selection (Full condition) for each domain-model pair.
  Claude shifts toward $T_3$ in later iterations on both domains.
  GPT diagnoses $T_1$ at every iteration on both domains.}
  \label{fig:stage-iter}
\end{figure*}

Table~\ref{tab:stage-diag} provides the per-domain stage selection breakdown.
The Claude $T_3$ concentration on Traffic intensifies across iterations: 41\,\% at iteration~1, 67\,\% at iteration~2, and 71\,\% at iteration~3 (Figure~\ref{fig:stage-iter}).
GPT's $T_1$ concentration on Traffic stays high throughout (84\,\%, 85\,\%, 79\,\%).
On Wildlife, GPT diagnoses $T_1$ in nearly every iteration.

\section{NLI Category Detail}
\label{sec:nli-appendix}

Table~\ref{tab:per-cat} reports the rate of each NLI category in the UNSAT and SAT pools, pooled across all 14 rows in Table~\ref{tab:drift-summary}.
Table~\ref{tab:cross-tab-full} gives the full breakdown.

\begin{table}[t]
\centering
\footnotesize
\setlength{\tabcolsep}{6pt}
\begin{tabular}{@{}lccc@{}}
\toprule
\textbf{NLI category} & \textbf{UNSAT pool} & \textbf{SAT pool} & \textbf{SAT $-$ UNSAT} \\
 & ($N{=}1275$) & ($N{=}310$) & \textbf{(pp)} \\
\midrule
Equivalent     & 42.4\% & 38.4\% & $-4.0$ \\
Related        & \phantom{0}9.0\% & \phantom{0}7.7\% & $-1.3$ \\
Strengthened   & 21.7\% & 15.5\% & $-6.2$ \\
Weakened       & 10.7\% & \phantom{0}9.7\% & $-1.0$ \\
Unrelated      & 10.0\% & 15.5\% & $+5.4$ \\
Contradiction  & \phantom{0}6.1\% & 13.2\% & $+7.1$ \\
\bottomrule
\end{tabular}
\caption{Per-category NLI rates in the UNSAT and SAT pools, pooled across all
14 rows in Table~\ref{tab:drift-summary} (3 repair methods $\times$ 4
model/domain pairs, plus 2 cross-model variants).
Negative values in the rightmost column mean the category is more frequent in UNSAT than in SAT.
Only Unrelated and Contradiction concentrate in SAT.
Strengthened concentrates in UNSAT, reflecting verbose legal back-translations of correctly formalized rules.
The other categories distribute approximately evenly.}
\label{tab:per-cat}
\end{table}

\begin{table*}[t]
\centering
\footnotesize
\setlength{\tabcolsep}{4pt}
\definecolor{cGood}{HTML}{C8E6C9}
\definecolor{cOk}{HTML}{DCEDC8}
\definecolor{cWarn}{HTML}{FFF9C4}
\definecolor{cBad}{HTML}{FFCDD2}
\begin{tabular}{@{}llccccccc@{}}
\toprule
\textbf{Domain / Model} & \textbf{Method / Verdict} & \textbf{Eq} & \textbf{Re} & \textbf{St} & \textbf{Wk} & \textbf{Un} & \textbf{Co} & \textbf{Total} \\
\midrule
\multirow{6}{*}{Traffic / Claude}
 & Random / UNSAT      & \cellcolor{cGood}55 & \cellcolor{cOk}7  & \cellcolor{cOk}18 & \cellcolor{cOk}6  & \cellcolor{cWarn}10 & \cellcolor{cBad}4  & 100 \\
 & Random / SAT        & 21 & 5 & 8  & 6 & 5  & 5  & 50 \\
 & Regenerate / UNSAT  & \cellcolor{cGood}56 & \cellcolor{cOk}7  & \cellcolor{cOk}17 & \cellcolor{cOk}5  & \cellcolor{cWarn}9  & \cellcolor{cBad}3  & 97 \\
 & Regenerate / SAT    & 28 & 1 & 9  & 4 & 4  & 7  & 53 \\
 & Full / UNSAT        & \cellcolor{cGood}73 & \cellcolor{cOk}13 & \cellcolor{cOk}24 & \cellcolor{cOk}6  & \cellcolor{cWarn}4  & \cellcolor{cBad}8  & 128 \\
 & Full / SAT          & 13 & 1 & 4  & 1 & 1  & 2  & 22 \\
\midrule
\multirow{8}{*}{Traffic / GPT}
 & Random / UNSAT      & \cellcolor{cGood}62 & \cellcolor{cOk}12 & \cellcolor{cOk}12 & \cellcolor{cOk}14 & \cellcolor{cWarn}9  & \cellcolor{cBad}9  & 118 \\
 & Random / SAT        & 12 & 3 & 4  & 4 & 6  & 3  & 32 \\
 & Regenerate / UNSAT  & \cellcolor{cGood}67 & \cellcolor{cOk}13 & \cellcolor{cOk}14 & \cellcolor{cOk}14 & \cellcolor{cWarn}12 & \cellcolor{cBad}7  & 127 \\
 & Regenerate / SAT    & 10 & 1 & 1  & 5 & 1  & 5  & 23 \\
 & Full / UNSAT        & \cellcolor{cGood}66 & \cellcolor{cOk}12 & \cellcolor{cOk}14 & \cellcolor{cOk}15 & \cellcolor{cWarn}9  & \cellcolor{cBad}8  & 124 \\
 & Full / SAT          & 10 & 2 & 4  & 1 & 5  & 4  & 26 \\
 & Full~(Cdx) / UNSAT  & \cellcolor{cGood}74 & \cellcolor{cOk}16 & \cellcolor{cOk}13 & \cellcolor{cOk}16 & \cellcolor{cWarn}8  & \cellcolor{cBad}9  & 136 \\
 & Full~(Cdx) / SAT    & 2  & 1 & 3  & 3 & 2  & 3  & 14 \\
\midrule
\multirow{6}{*}{Wildlife / Claude}
 & Random / UNSAT      & \cellcolor{cGood}12 & \cellcolor{cOk}6  & \cellcolor{cOk}20 & \cellcolor{cOk}7  & \cellcolor{cWarn}12 & \cellcolor{cBad}3  & 60 \\
 & Random / SAT        & 6  & 2 & 2  & 1 & 5  & 1  & 17 \\
 & Regenerate / UNSAT  & \cellcolor{cGood}13 & \cellcolor{cOk}5  & \cellcolor{cOk}24 & \cellcolor{cOk}5  & \cellcolor{cWarn}12 & \cellcolor{cBad}4  & 63 \\
 & Regenerate / SAT    & 4  & 2 & 2  & 2 & 2  & 2  & 14 \\
 & Full / UNSAT        & \cellcolor{cGood}16 & \cellcolor{cOk}7  & \cellcolor{cOk}21 & \cellcolor{cOk}6  & \cellcolor{cWarn}14 & \cellcolor{cBad}2  & 66 \\
 & Full / SAT          & 2  & 0 & 2  & 2 & 3  & 2  & 11 \\
\midrule
\multirow{8}{*}{Wildlife / GPT}
 & Random / UNSAT      & \cellcolor{cGood}10 & \cellcolor{cOk}3  & \cellcolor{cOk}24 & \cellcolor{cOk}11 & \cellcolor{cWarn}6  & \cellcolor{cBad}5  & 59 \\
 & Random / SAT        & 5  & 2 & 3  & 1 & 3  & 3  & 17 \\
 & Regenerate / UNSAT  & \cellcolor{cGood}13 & \cellcolor{cOk}4  & \cellcolor{cOk}25 & \cellcolor{cOk}10 & \cellcolor{cWarn}7  & \cellcolor{cBad}6  & 65 \\
 & Regenerate / SAT    & 4  & 1 & 0  & 0 & 5  & 1  & 11 \\
 & Full / UNSAT        & \cellcolor{cGood}11 & \cellcolor{cOk}3  & \cellcolor{cOk}24 & \cellcolor{cOk}10 & \cellcolor{cWarn}7  & \cellcolor{cBad}5  & 60 \\
 & Full / SAT          & 2  & 3 & 5  & 0 & 4  & 2  & 16 \\
 & Full~(Cdx) / UNSAT  & \cellcolor{cGood}13 & \cellcolor{cOk}7  & \cellcolor{cOk}27 & \cellcolor{cOk}11 & \cellcolor{cWarn}9  & \cellcolor{cBad}5  & 72 \\
 & Full~(Cdx) / SAT    & 0  & 0 & 1  & 0 & 2  & 1  & \phantom{0}4 \\
\bottomrule
\end{tabular}
\caption{Full 6-category NLI cross-tabulations partitioned by SMT verdict, for
 all 14 rows in Table~\ref{tab:drift-summary} (3 repair methods $\times$ 4
 domain-model pairs, plus 2 cross-model variants).
Eq = Equivalent, Re = Related, St = Strengthened, Wk = Weakened, Un = Unrelated, Co = Contradiction.
In UNSAT rows, shading indicates semantic quality:
\colorbox{cGood}{\strut\,faithful\,},
\colorbox{cOk}{\strut\,partial drift\,},
\colorbox{cWarn}{\strut\,unrelated\,},
\colorbox{cBad}{\strut\,contradiction\,}.
SAT rows are uncolored.
Seven rules across four domain-model pairs returned unknown or errored on the final equivalence check.
They are counted in the SAT pool because they were not verified equivalent.}
\label{tab:cross-tab-full}
\end{table*}

\FloatBarrier
\raggedbottom

\section{Drift Before Repair}
\label{sec:pre-repair}

The drift split in \S\ref{subsec:nli-results} is measured after repair, so it could reflect repair
leaving the harder rules unsolved rather than the verdict itself.
Table~\ref{tab:pre-repair} measures the same split on the baseline pipeline,
before any repair runs.
Observe that the gap is already present before the repair.

\begin{table}[t]
\centering
\scriptsize
\setlength{\tabcolsep}{4pt}
\begin{tabular}{@{}lccc@{}}
\toprule
\textbf{Domain / Model} & \textbf{UNSAT drift} & \textbf{SAT drift} & \textbf{Ratio} \\
\midrule
Traffic / Claude  & \phantom{0}7.5\% (5/67)  & 15.7\% (13/83) & 2.10 \\
Traffic / GPT     & 13.0\% (12/92) & 29.3\% (17/58) & 2.25 \\
Wildlife / Claude & 25.0\% (12/48) & 34.5\% (10/29) & 1.38 \\
Wildlife / GPT    & 19.6\% (10/51) & 28.0\% (7/25)  & 1.43 \\
\midrule
\textbf{Pooled}   & \textbf{15.1\% (39/258)} & \textbf{24.1\% (47/195)} & \textbf{1.59} \\
\bottomrule
\end{tabular}
\caption{Drift rate on the baseline pipeline, split by the initial equivalence verdict, with rule
counts in parentheses.
Pool sizes match the \textbf{None} column of Table~\ref{tab:repair-summary}.
The Wildlife / GPT pools total 76 because one rule yielded no Stage-1 encoding.}
\label{tab:pre-repair}
\end{table}

\section{Human Evaluation}
\label{sec:human-eval}

Both signals in this paper are automatic, so neither replaces human judgment.
We ran two small studies to anchor them.

\paragraph{Faithfulness.}
We drew a sample of 25 rules that spans both rules whose roundtrip passed the equivalence check and rules whose roundtrip failed it.
The sample comes from the two Claude runs and is split across the two domains: 8 passing and 6 failing rules from Traffic, 6 passing and 5 failing from Wildlife.
For each rule the annotator saw only the original rule and its formalization, not the solver's verdict, and judged whether the formalization captures the meaning of the rule.
The annotator is an author with formal-methods training.
Table~\ref{tab:human-faith} reports the result.
The judgments line up with the verdict.
Formalizations of passing rules were judged not faithful in 1 of 14 cases (7\,\%), against 3 of 11 (27\,\%) for failing rules.

\begin{table}[t]
\centering
\scriptsize
\setlength{\tabcolsep}{5pt}
\begin{tabular}{@{}lcccc@{}}
\toprule
\textbf{Verdict} & \textbf{Rules} & \textbf{Faithful} & \textbf{Minor} & \textbf{Unfaithful} \\
\midrule
UNSAT (equivalent)   & 14 & 13 (93\%) & 1 (7\%) & 0 \\
SAT (not equivalent) & 11 & \phantom{0}8 (73\%) & 3 (27\%) & 0 \\
\bottomrule
\end{tabular}
\caption{Human faithfulness judgments, split by the roundtrip equivalence verdict.
\emph{Faithful} means the formalization captures the meaning of the rule.
\emph{Minor} means mostly right with one detail off, such as a bound, a qualifier, or a scope.
\emph{Unfaithful} means the formalization misses or distorts the rule.
\emph{Not faithful} in the text means \emph{Minor} or \emph{Unfaithful}.
The annotator judged meaning rather than style and did not see the verdict.}
\label{tab:human-faith}
\end{table}

\paragraph{Diagnosis.}
On the 11 rules whose roundtrip failed, the annotator read the four roundtrip artifacts and marked which stage first lost meaning.
The stage the system had diagnosed was not shown.
The two agreed in 7 of the 11 cases.

Both studies use a single annotator on a small sample from the Claude pipeline.
We read them as a directional anchor rather than an exact rate.

\section{Run-to-Run Variance}
\label{sec:variance}

The pipeline exposes no seed, so a replicate is an independent end-to-end re-execution at the same
settings: temperature 0.3, top-$p$ 0.85, byte-identical prompts and schema, and the same model
identifiers.
We ran two replicates for each of the four domain-model pairs, eight runs in total, each covering
baseline generation and all three repair methods.
Table~\ref{tab:var-unsat} reports final UNSAT counts alongside the single run reported in the main text.

\begin{table}[t]
\centering
\scriptsize
\setlength{\tabcolsep}{4pt}
\begin{tabular}{@{}llcccc@{}}
\toprule
\textbf{Domain/Model} & \textbf{Method} & \textbf{Paper} & \textbf{Rep 1} & \textbf{Rep 2} & \textbf{Mean $\pm$ std} \\
\midrule
\multirow{3}{*}{Traffic / Claude}
 & Full       & 128 & 125 & 132 & 128.5 $\pm$ 3.5 \\
 & Random     & 100 & 102 & 108 & 105.0 $\pm$ 3.0 \\
 & Regenerate & \phantom{0}97 & \phantom{0}97 & 104 & 100.5 $\pm$ 3.5 \\
\midrule
\multirow{3}{*}{Traffic / GPT}
 & Full       & 124 & 122 & 126 & 124.0 $\pm$ 2.0 \\
 & Random     & 118 & 124 & 126 & 125.0 $\pm$ 1.0 \\
 & Regenerate & 127 & 126 & 130 & 128.0 $\pm$ 2.0 \\
\midrule
\multirow{3}{*}{Wildlife / Claude}
 & Full       & 66 & 65 & 68 & 66.5 $\pm$ 1.5 \\
 & Random     & 60 & 58 & 63 & 60.5 $\pm$ 2.5 \\
 & Regenerate & 63 & 64 & 69 & 66.5 $\pm$ 2.5 \\
\midrule
\multirow{3}{*}{Wildlife / GPT}
 & Full       & 60 & 61 & 60 & 60.5 $\pm$ 0.5 \\
 & Random     & 59 & 63 & 58 & 60.5 $\pm$ 2.5 \\
 & Regenerate & 65 & 65 & 65 & 65.0 $\pm$ 0.0 \\
\bottomrule
\end{tabular}
\caption{Final UNSAT count per replicate.
\textbf{Paper} is the run reported in Table~\ref{tab:repair-summary}.
Mean and std are over the two replicates.
Under Full, every value reported in the main text falls inside its replicate
range for all four domain-model pairs.}
\label{tab:var-unsat}
\end{table}

The counts are stable.
Under Full, every value reported in the main text falls inside its replicate
range for all four domain-model pairs,
and Regenerate at least matches Full for GPT in both domains, reproducing the diagnosis-collapse
effect of \S\ref{subsec:gpt-collapse}.
One method comparison does not reproduce: Full's lead over Regenerate on UNSAT count disappears for
Wildlife / Claude, with both at 66.5, though Full still beats Random there and keeps its cost
advantage.

\section{Data Availability}
\label{sec:data-availability}

We evaluate on two public statutory corpora, both in English.
The traffic corpus consists of 150 rules derived from the Texas Transportation Code, a public-domain statutory text published by the Texas Legislature.\footnote{\url{https://statutes.capitol.texas.gov/Docs/TN/htm/TN.545.htm}}
The wildlife corpus consists of 77 rules derived from the Texas Parks and Wildlife Code, also published by the Texas Legislature.\footnote{\url{https://statutes.capitol.texas.gov/?link=PW}}
As U.S.\ state-government-authored statutes, both source texts are not subject to copyright.
The source corpora contain no personally identifiable information or offensive content, as they are statutory text describing legal subjects in the abstract.

Code, data, prompts, and domain schemas are publicly available.\footnote{\url{https://github.com/daneshvar-amrollahi/roundtrip-autoformalization}}
External artifacts are used under their respective open-source licenses: Z3 (MIT), BART weights (MIT), and MultiNLI (mixed permissive licenses per source).
All external artifacts (Z3, BART-large-mnli, MultiNLI, Texas statutes, and commercial LLM APIs) are used consistently with their intended purposes as solvers, inference models, training data, public reference text, and language-generation services, respectively.
Our released artifacts (code, data, prompts, schemas) are intended for research use on autoformalization and roundtrip-style verification.

\end{document}